\documentclass[runningheads]{llncs}

\usepackage{graphicx}
\usepackage{amsmath} 
\usepackage{amssymb}  
\usepackage{diagbox}
\usepackage{xcolor} 

\usepackage{subcaption}

\hypersetup{
    colorlinks=true,
    linkcolor=blue,
    filecolor=blue,      
    urlcolor=blue,
    citecolor=blue,
    bookmarks=true,
}
\begin{document}
\title{Auto-labelling of Markers in Optical Motion Capture by Permutation Learning}
\titlerunning{Auto-labelling of Markers in Optical MoCap by Permutation Learning}
%
\author{Saeed Ghorbani\inst{1}\orcidID{0000-0002-3227-9013} \and
Ali Etemad\inst{2}\orcidID{0000-0001-7128-0220} \and
Nikolaus F. Troje\inst{1}\orcidID{0000-0002-1533-2847}}
\authorrunning{S. Ghorbani et al.}
%
\institute{York University, Toronto ON, Canada\\
\email{{saeed@eecs.yorku.ca, troje@yorku.ca}}\\
\and
Queen's University, Kingston ON, Canada\\
\email{ali.etemad@queensu.ca}}
\maketitle              
\raggedbottom
\begin{abstract}\label{section: Abstract}
Optical marker-based motion capture is a vital tool in applications such as motion and behavioural analysis, animation, and biomechanics. Labelling, that is, assigning optical markers to the pre-defined positions on the body, is a time consuming and labour intensive post-processing part of current motion capture pipelines. The problem can be considered as a ranking process in which markers shuffled by an unknown permutation matrix are sorted to recover the correct order. In this paper, we present a framework for automatic marker labelling which first estimates a permutation matrix for each individual frame using a differentiable permutation learning model and then utilizes temporal consistency to identify and correct remaining labelling errors. Experiments conducted on the test data show the effectiveness of our framework. 
\end{abstract}
\section{Introduction}\label{section: Introduction}
Optical motion capture is an important technology for obtaining high accuracy human body and motion information. Motion capture has been widely used in applications such as human motion analysis \cite{Troje2008,Etemad2016}, producing realistic character animation \cite{Loper2014,Pons2015}, and validation of computer vision and robotic tasks \cite{Sigal2010,Ionescu2014}
During the recording step, the motion of passively reflecting markers, which are attached to the body according to a predefined marker layout, are tracked by multiple high-resolution (spatial and temporal) infrared cameras. Then, the 3D positions of markers are computed from 2D data recorded by each of the calibrated cameras by means of triangulation. The result of this process is a set of 3D trajectories listed in random order. Each trajectory represents the motion of a single marker in terms of its 3D position over time. The first step after recording phase is to label each trajectory, therefore assigning it to a specific body location. This process can be very time-consuming and therefore a friction point for many motion capture systems. Furthermore, this process is susceptible to user errors. Labelling is more challenging when one or multiple markers are occluded (for simplicity we use term \emph{occlusion} for any type of missing marker in the data) over time, hence splitting the motion of these markers into multiple trajectories. Commercial motion capture softwares have designed frameworks to reduce the amount of manual work in this step. In these frameworks, a model can learn the geometry of that participant's body and applies it to label subsequent measurements. However, they typically require manual initialization where one or more motion capture sequences have to be labelled manually for each participant. 
Here, we are proposing an end-to-end, data-driven approach for automatic labelling which does not require a manual initialization. We formulated the labelling at each frame as a permutation estimation where shuffled markers are ranked based on a pre-defined order. Then, in a trajectory labelling step, temporal consistency is used to correct mislabelled markers. Our framework can be reliably run in real time that potentially results in a faster, cheaper, and more consistent data processing in motion capture pipelines. 
\section{Related Works}\label{section: Related Works}
Typically, motion capture labelling comprises of two main steps: the initialization step, where the initial correspondences are established for the first frame, and the tracking step which can be defined as keeping track of labelled markers in the presence of occlusion, ghost markers, and noise. 
A number of approaches have been introduced to address the tracking of manually initialized motion capture data. Holden \cite{Holden2018} proposed a deep de-noising feed-forward network that outputs the joint locations directly from corrupted markers. Herda et al. \cite{Herda2001} proposed an approach to increase the robustness of optical markers specifically during visibility constraints and occlusions by using the kinematics information provided by a generic human skeletal model. Yu et al. \cite{Yu2007} proposed an online motion capture approach for multiple subjects which also recovers missing markers. They used the standard deviation of the distance between each pair of markers to cluster the markers into a number of rigid bodies. Having fitted rigid bodies, they labelled the markers using a structural model for each rigid body and a motion model for each marker. Loper \cite{Loper2014} proposed a marker placement refinement by optimizing the parameters of a statistical body model in a generative inference process.
Another group of approaches attempt to minimize user intervention by automating the initialization step as well. Holzreiter \cite{Holzreiter2005} trained a neural network to estimate the positions of sorted markers from a shuffled set. Labelling of the markers was carried out by pairing up the estimated marker locations with the shuffled set using the nearest neighbour search. Meyer et al. \cite{Meyer2014} estimated the skeletal configuration by least-squares optimization and exploited the skeletal model to automatically label the markers. They applied the Hungarian method for optimal assignment of observation to markers while requiring each subject to go into a T-pose to initialize the skeletal tracker.
Schubert et al. \cite{Schubert2015} improved their approach by designing a pose-free initialization step, searching over a large database of poses. 
Finally, Han et al. \cite{Han2018}  and Maycock et al. \cite{Maycock2015} proposed auto-labelling approaches specifically designed for hands. Han et al. \cite{Han2018} proposed a technique to label the hand markers by formulating the task as a keypoint regression problem. They rendered the marker locations as a depth image and fed it into a convolutional neural network which outputs the labelled estimated 3D locations of the markers. Then, they used a bipartite matching method to map the labels onto the actual 3D markers. Maycock. et. al \cite{Maycock2015} used inverse kinematics to filter out unrealistic postures and computed the assignment between model nodes and 3D points using an adapted version of the Hungarian method. Although their approach was focused on hands, it can be applied to human body motion. 
\section{System Overview}\label{section: SystemOverview}
\begin{figure}[t]
    \hspace*{0.05 \textwidth}\includegraphics[width=0.9 \textwidth]{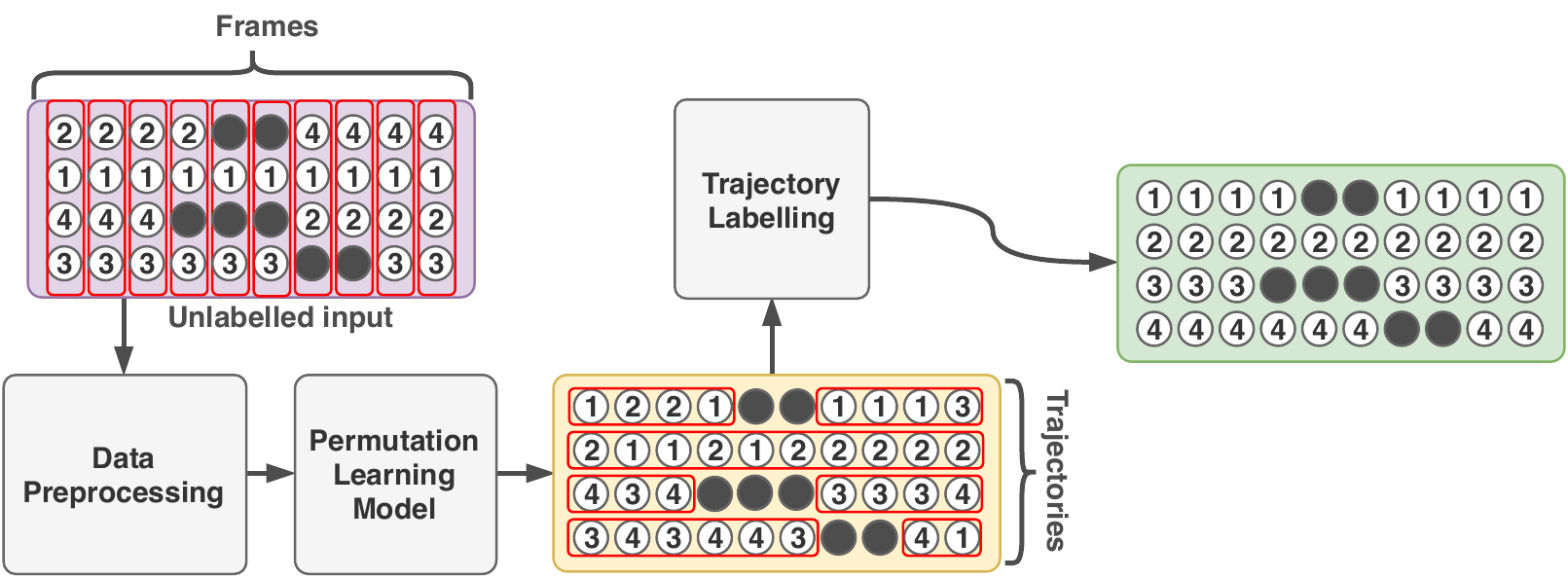}
    \caption{An overview of our proposed framework: the input to our system is a sequence of unlabelled (shuffled) 3D trajectories. After the preprocessing stage, the label for each marker is estimated by applying permutation learning to each individual frame. The resulting labelled frames are then concatenated to form trajectories again. These trajectories are then used as input to the trajectory labelling stage where a temporal consistency constraint is used to correct mislabelled markers}
    \label{figure: Main}
\end{figure}

The first stage of our framework is a preprocessing step which is applied on individual frames, making the array of markers invariant to spatial transformations. For the next stage, we propose a data-driven approach that avoids the need for manual initialization by formulating automatic labelling as a permutation learning problem for each individual frame. Towards this end, we present a permutation learning model which can be trained end-to-end using a gradient-based optimizer. We exploited the idea of relaxing our objective function by using doubly-stochastic matrices as a continuous approximation of permutation matrices \cite{Adams2011}. During the running phase, each individual frame is automatically labelled using the proposed permutation learning model. The result is a sequence of individually labelled frames where each trajectory might be assigned to multiple labels over time. We then evaluate temporal consistency in the resulting trajectories and use it to identify and correct the labelling errors that occurred during the previous stage. To correct the inconsistencies in each marker trajectory, a score is computed for each candidate label using a confidence-based score function. Then, the label with the highest score will be assigned to the marker trajectory in a winner-takes-all scheme. Fig. \ref{figure: Main} shows the block diagram of our main framework.
\subsection{Data Preprocessing} \label{section: Data Preprocessing}
Prior to applying our permutation learning model, we ensure that the input data are invariant to translation, orientation, and the size of the subjects. We first calculated the centroid of the array of markers for each frame and then subtracted from the marker locations to make each frame invariant to translation. To make the data invariant to the orientation of the subject, we applied principal component analysis (PCA) to the cloud of the markers. We first aligned the direction with the largest principal component with the $z$-axis. Then, the second principal component was aligned to the $x$-axis to make the poses invariant to the rotations around $z$-axis. Finally, the size of the subject was normalized by scaling the values between $ 0 $ and $ 1 $ independently for each of the three spatial dimensions. 
\subsection{Permutation Learning Model}\label{subsection: Permutation Learning Model}
The motion capture data at each frame can be represented by the 3D positions of $N$ markers utilized in the recording process. \emph{Labelling} is defined as assigning the 3D positions of these markers to specific, fixed body locations. This process can be described as permuting a set of shuffled 3D elements to match a pre-defined order. Let us define a labelled frame as $X = \left[ m_{1},\ m_{2},\dots,\ m_{N} \right]^\top,$  to be an ordered array of $N$ markers, where $m_{i}$ represents the 3D position of the $i_{th}$ marker.  Then, a shuffled frame $\tilde{X}=PX$ can be considered as a permuted version of $X$ where the markers are permuted by a permutation matrix $P \in \{0,\ 1\}^{N\times N}$. Hence, given a shuffled frame, the original frame can be recovered by multiplying the shuffled version with the inverse of the respective permutation matrix. It should be noted that for a permutation matrix $P$, $P^{\top}=P^{-1}$. Our goal in this step is to design a trainable parameterized model $f_{ \theta }:\mathcal{ X }^{ N } \rightarrow \mathcal{ S }_N$ which takes a shuffled frame $\tilde{X}$ as input and estimates the permutation matrix $P$ that was originally applied to the frame. Then, having the permutation matrix $P$ we can recover the sorted frame $X$, $X=P^{\top}\tilde{X}$.

The main difficulty in training a permutation learning model using backpropagation is that the space of permutations is not continuous which prohibits computation of the gradient of the objective function with respect to the learning parameters since it is not differentiable in terms of the permutation matrix elements. To address this problem, Adams et al. \cite{Adams2011} proposed the idea of utilizing a continuous distribution over assignments by using doubly-stochastic matrices as differentiable relaxations of permutation matrices. This approach has been successfully exploited in other applications \cite{santa2017,rezatofighi2018}. A DSM is a square matrix populated with non-negative real numbers where each of the rows and columns sums to $1$. All $N\times N$ DSM matrices form a convex polytope known as the Birkhoff polytope $\mathcal{B}_N$ lying on a $(N-1)^2$ dimensional space where the set of all $N\times N$ permutation matrices are located exactly on the vertices of this polytope \cite{Birkhoff1946}. Therefore, DSMs can be considered as continuous relaxations of corresponding permutation matrices. We can interpret each column $i$ of a DSM as a probability distribution over labels to be assigned to the $i_{th}$ marker. Also, all rows summing to $1$ ensures the inherent structure of permutation matrices. Accordingly, instead of mapping directly from 3D positions to the permutation matrices, we propose to learn a model $g_{ \theta }:\mathcal{ X }^{ N } \rightarrow \mathcal{ W }_N$, where ${ \mathcal{ W } }_N$ is the set of all $N\times N$ DSMs. That way, computing the permutation matrix from the DSM simply becomes a bipartite matching problem. \par
\begin{figure}[t]
    \hspace*{0.03 \textwidth}\includegraphics[width=0.9 \textwidth]{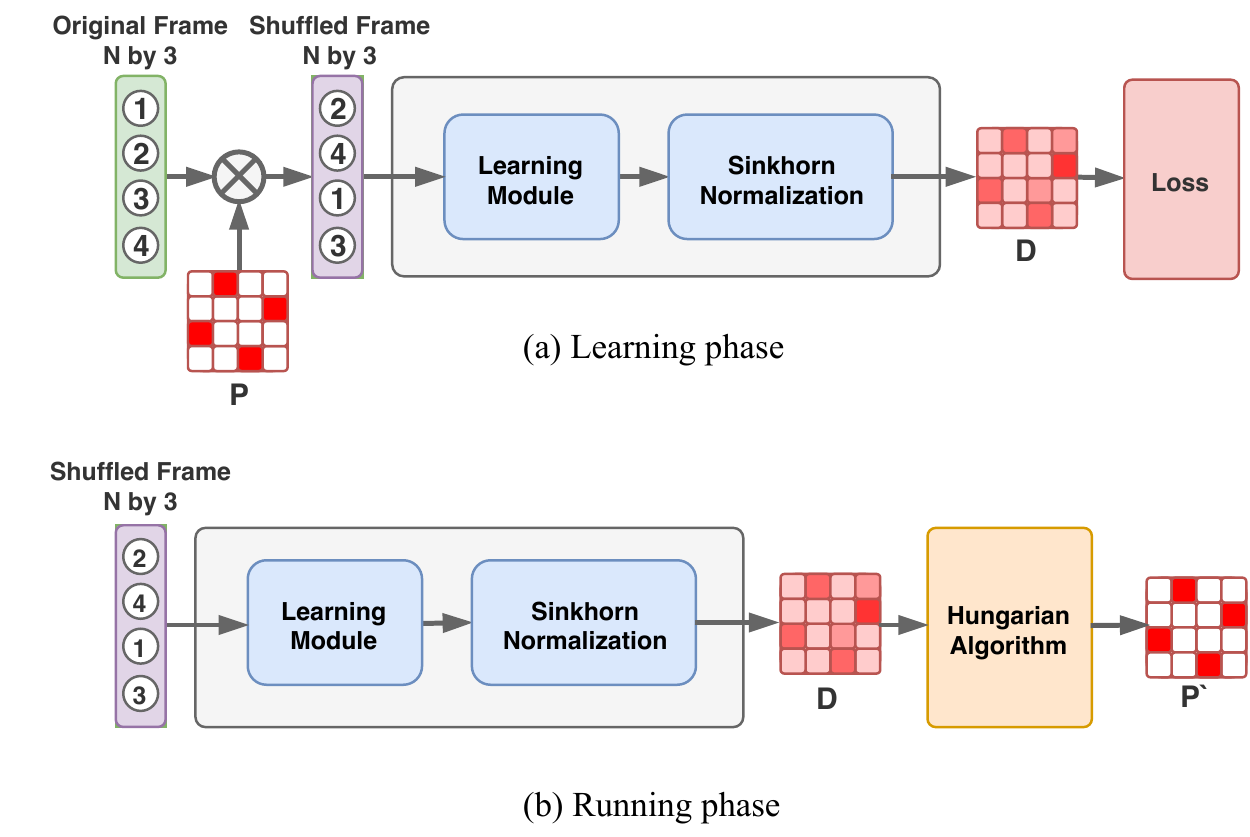}
    \caption{An overview of our permutation learning model: In (a), the learning phase is depicted, where the parameters of the learning module are optimized to minimize the cross-entropy loss. In (b), the running phase is illustrated, where the Hungarian algorithm is applied to the outputted DSM to estimate the optimal permutation matrix.}
    \label{figure: LearningModel}
\end{figure} 
Our permutation learning model is composed of two main modules: the learning module and the Sinkhorn normalization (see Fig. \ref{figure: LearningModel}). The learning module $h_{ \theta }:\mathcal{ X }^{ N } \rightarrow \mathbb{ R }_{+}^{N\times N}$ is the parameterized component of our model, which learns the feature representation of the data structure from the available poses, taking the 3D positions in each frame and outputting an unconstrained square matrix. We implement this module as a feed-forward deep residual neural network \cite{He2016}. The last dense layer consists of $N\times N$ nodes with a sigmoid activation outputting a $N\times N$ non-negative matrix. The learning module is illustrated in Fig. \ref{figure: LearningModel}. One na\"ive approach would be to treat the problem as a multi-class, multi-label classification task with $N^2$ classes. However, this approach would ignore the inherent structure of permutation matrices with the possibility of resulting in impossible solutions. To enforce the optimizer to avoid these erroneous solutions we need our model to output a DSM which replicates the inherent structure of permutation matrices. An effective way to convert any unconstrained non-negative matrix to a DSM is using an iterative operator known as Sinkhorn normalization \cite{Sinkhorn1964,Adams2011}. This method normalizes the rows and columns iteratively where each pair of iteration is defined as:
\begin{equation}
    S^{i}(M)=
    \begin{cases}
    M  \quad\quad  &\text{if $i=0$}\\
    \mathcal{T}_{R}(\mathcal{T}_{C}(S^{i-1}(M)))	&\text{otherwise,}
    \end{cases}
    \label{equation: Equation1}
\end{equation}
where $\mathcal{T}_{R}^{i,j}(M) = \frac{M_{i,j}}{\sum_{k}M_{i,k}}$ and $\mathcal{T}_{C}^{i,j}(M) = \frac{M_{i,j}}{\sum_{k}M_{k,j}}$ are the row and column-wise normalization operators, respectively.
The Sinkhorn normalization operator is defined as $S^{\infty}:\mathbb{ R }_{+}^{N\times N} \rightarrow \mathcal{ W }_N $ where the output converges to a DSM. We approximate the Sinkhorn normalization by an incomplete version of it with $i<\infty$ pairs of iteration. The defined Sinkhorn normalization function is differentiable and the gradients of the learning objective can be computed by backpropagating through the unfolded sequence of row and column-wise normalizations, unconstrained matrix, and finally learning module parameters. 

During the running phase, a single permutation matrix $P^{'}$ must be predicted by finding the closest polytope vertex to the doubly-stochastic matrix $D$ produced by the model. This can be formulated as a bipartite matching problem where the cost matrix is $C = 1 - D$. As a result, we use the Hungarian algorithm over the cost matrix to find the optimal solution of the matching problem. Finally, each individual frame is sorted (labelled) by the transpose of the corresponding estimated permutation. The learning and running phases are illustrated in Fig. \ref{figure: LearningModel}.
\subsection{Trajectory Labelling}\label{sebsection: Trajectory Labelling}
After applying our permutation learning model to all the frames in the entire sequence, we have a sequence of individually labelled frames ordered in time. However, the integration of sequences of individual marker locations into trajectories that extend over time, which is already conducted by the motion capture system during data collection, and the expectation that labels should remain constant during the motion trajectory, can be used to enforce temporal consistency. Each trajectory can be defined as the sequence of tracked marker locations which ends with a gap or when recording stops. Therefore, in each motion sample, the movement of each marker might be presented in multiple trajectories over time. We can exploit the temporal consistency of each trajectory to correct the wrong predictions for each marker during the trajectory. One na\"ive idea is to assign each trajectory to the label with the highest number of votes in the assignment predictions for the corresponding marker. However, there are situations where a label has been assigned to a marker with the highest number of times but with low confidence. Thus, we propose a winner-takes-all approach where a score is computed for each label which has been assigned at least once to the marker in the trajectory. Then the winner with the highest score will be assigned to the corresponding trajectory. The score for label $ i $ assigned to the query marker $j$ is computed as follows:
\begin{equation}
    S_{ i } = { |T_{ i }| } ^ { q } \left( \sum_{ t \in T_{ i } } |c_{ i, j }^{(t)}|^{ p } \right) ^ \frac{ 1 }{ p }
    \label{equation: Equation2}
\end{equation}
where $T_{ i }$ is the set of frame indices at which label $ i $ has been assigned to the query marker during its trajectory. $|T_{ i }|$ is the cardinality of $T_{ i }$. $ p $ and $ q $ are hyperparameters which are chosen during validation step. The second term for $p=0$ is defined as $|T_{ i }|$. $c_{ i, j }^{(t)}$ represents the degree of confidence for assigning the label $ i $ to the marker $j$ at frame $t$. The details of $c_{ i, j }^{(t)}$ formulation are discussed in section \ref{equation: Degree of Confidence}. 
\subsection{Degree of Confidence}\label{equation: Degree of Confidence}
During the running phase, each column of the outputted DSM matrix represents a distribution over labels for the corresponding marker which can be interpreted as the model's belief in each label. When the model is confident in labelling all markers, the estimated DSM matrix is close to the true permutation matrix on the polytope surface and all of these distributions peak at the true label. On the other hand, when a marker is hard to label, the corresponding distribution might not have a sharp peak at the true label. We compute a degree of confidence for each predicted label $i$ assigned to marker $j$ at frame $t$ as the distance between the model's belief for label $i$ and the highest belief, as follows:
\begin{equation}
    c_{ i, j }^{(t)} = D_{ i,j }^{(t)} - \max_{\scriptstyle 1 \leq k \leq N \atop \scriptstyle k \neq i }{D_{k,j}^{(t)}},
    \label{equation: Equation3}
\end{equation}
where $D^{(t)}$ is the DSM matrix produced by Sinkhorn normalization at frame $ t $. Note that the defined $c_{ i, j }^{(t)}$ can be negative in the situations where the assigned label and the index of maximum value in the distibution are not the same. Therefore, we use a min-max normalized version of it $(c_{ i, t } \leftarrow \frac{c_{ i, t } + 1}{2})$ in equation \ref{equation: Equation2} to make sure its range is between 0 and 1.
\section{Experiments and Evaluations} \label{section: Experiments and Evaluations}
\subsection{Data} \label{subsection: Data}
To evaluate our method we used a subset of Biomotion dataset \cite{Troje2008} recorded from $ 115 $ individuals at $ 120 $ frames per second using Vicon system. The subset includes four types of actions namely walking, jogging, sitting, and jumping, where each recording contains the 3D trajectories of $41$ markers. Some actions where recorded multiple times from the same individual. On average, we used $ 11.5 $ sequences per subject for a total of $ 1329 $ sequences. The total number of frames in our dataset is around $630$k frames. Data from $ 69 $ individuals were used for training, while the data from $ 46 $ different individuals were held out for testing and validation ($ 23 $ each). 
\subsection{Training} \label{subsection: Training}
\begin{figure}[t]
    \hspace*{0.1 \textwidth}
	\includegraphics[width=.40\paperwidth]{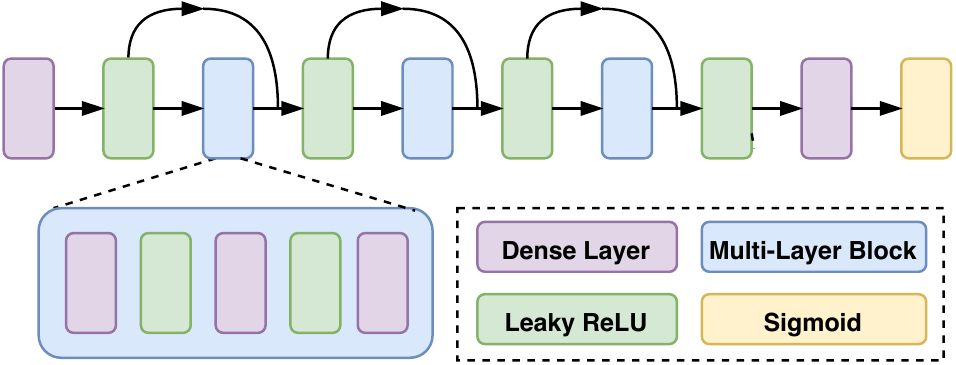}
	\caption{Schematic diagram of the feed-forward deep residual neural network used for permutation learning.}
	\label{figure: ResModel}
\end{figure}
The feed-forward residual network implemented as our learning module was designed with three residual blocks where each block contains three dense layers followed by a Leaky ReLU activation (see Fig. \ref{figure: ResModel}). The residual connections showd a smoothing behavior on our optimization landscape. Hyper-parameters of the network were chosen using a random search scheme \cite{Bergstra2012}.  

The original training set was constructed by applying $16$ random permutations on each of the $378$k training frames resulting in around $6.4$ million shuffled training frames. 
Then, to augment the training data with the occlusions, up to $5$ markers in each generated shuffled frame were randomly occluded by replacing the 3D position values by the center location $(0.5, 0.5, 0.5)$. Both the numbers of occlusions and the index of occluded markers were drawn from uniform distributions.

For training the model, we used an Adam optimizer with batches of size $ 32 $. The learning rate was initially set to $5\times10^{-5}$ and was reduced by a factor of $ 2 $ after each epoch when the validation loss increased. Our model was trained for $ 100 $ epochs using a cross-entropy loss function. The number of Sinkhorn iterations was set to $5$ since for each additional iteration the improvement in performance was very small while the running time was increased linearly. By performing these $5$ Sinkhorn iterations on our unconstrained matrix, the sum of squared distances between  $1$ and the row- and column-wise sums of the resulting matrix was less than $10^{-17}$.
\subsection{Permutation Learning Model Evaluation} \label{subsection: Permutation Learning Model Evaluation}
To evaluate our permutation learning model, we synthesized an evaluation set by applying $16$ random permutations followed by randomly introducing $0$ to $5$ occlusions into each frame of the test set. Table \ref{tab:table1} shows the performance of our model in different setups and compares them with the initialization steps proposed by Holzreiter et al. \cite{Holzreiter2005} and Maycock et al. \cite{Maycock2015}. First and second rows in the table \ref{tab:table1} show the accuracy results when the model was trained on the original training set and the occlusion augmented set, respectively. As anticipated, introducing occlusions into the training data improves the results for the test frames with occluded markers, which is usual in real scenarios. On the other hand, when the model is trained without occlusions the performance on occluded frames significantly decreases. \par
To evaluate the influence of Sinkhorn normalization, we replaced the Sinkhorn layer with a Softmax function over the rows of outputted DSM matrix and trained the parameters from scratch. The results for this setups are illustrated in the third row of table \ref{tab:table1}. Without Sinkhorn normalization, the output matrix ignores the inherent structure of permutation matrices resulting in a drop in the labelling performance.

\begin{table}[t]
    \centering
    \caption{A comparison of performance of different models in labelling a single test frame as an initialization step in the presence of a varying number of occlusions in the test frames (show in each column). The first and second rows illustrates the performance when the model is trained with and without occlusions, respectively. Third row, shows the results when the Sinkhorn layer is replaced by a Softmax function. It can be seen that the model trained on occlusion augmented training set outperforms the rest when having occlusions in the frames.}
    \vspace{0.2cm}
    \label{tab:table1}
    \begin{tabular}{|c|c|c|c|c|c|c|}
    \hline
    \diagbox[ dir=NW,innerleftsep=.1cm,innerrightsep=5pt]{Method}{\# Occs} & 0 & 1 & 2 & 3 & 4 & 5\\
    \hline \hline
    Ours + Occs &  97.11\% & \bfseries{96.56}\% & \bfseries{96.13}\%  & \bfseries{95.87}\% & \bfseries{95.75}\% & \bfseries{94.9}\%\\ \hline
    Ours + No Occs & \bfseries{98.72}\% & 94.41\% & 92.15\% & 88.75\% & 86.54\% & 85.0\% \\ \hline
    Ours w/o SN & 94.03\% & 91.12\% & 88.1\% & 84.27\% & 81.62\% & 77.78\% \\ \hline
    Maycock et al. & 83.18\% & 79.35\% & 76.44\% & 74.91\% & 71.17\% & 65.83\%\\ \hline
    Holzreiter et al. & 88.16\% & 79.0\% & 72.42\% & 67.16\% & 61.31\% & 52.1\%\\ \hline
    \end{tabular}
\end{table}
\begin{figure}[t]
    \centering
    \begin{subfigure}{.45\textwidth}
      \centering
      \includegraphics[width=1\linewidth]{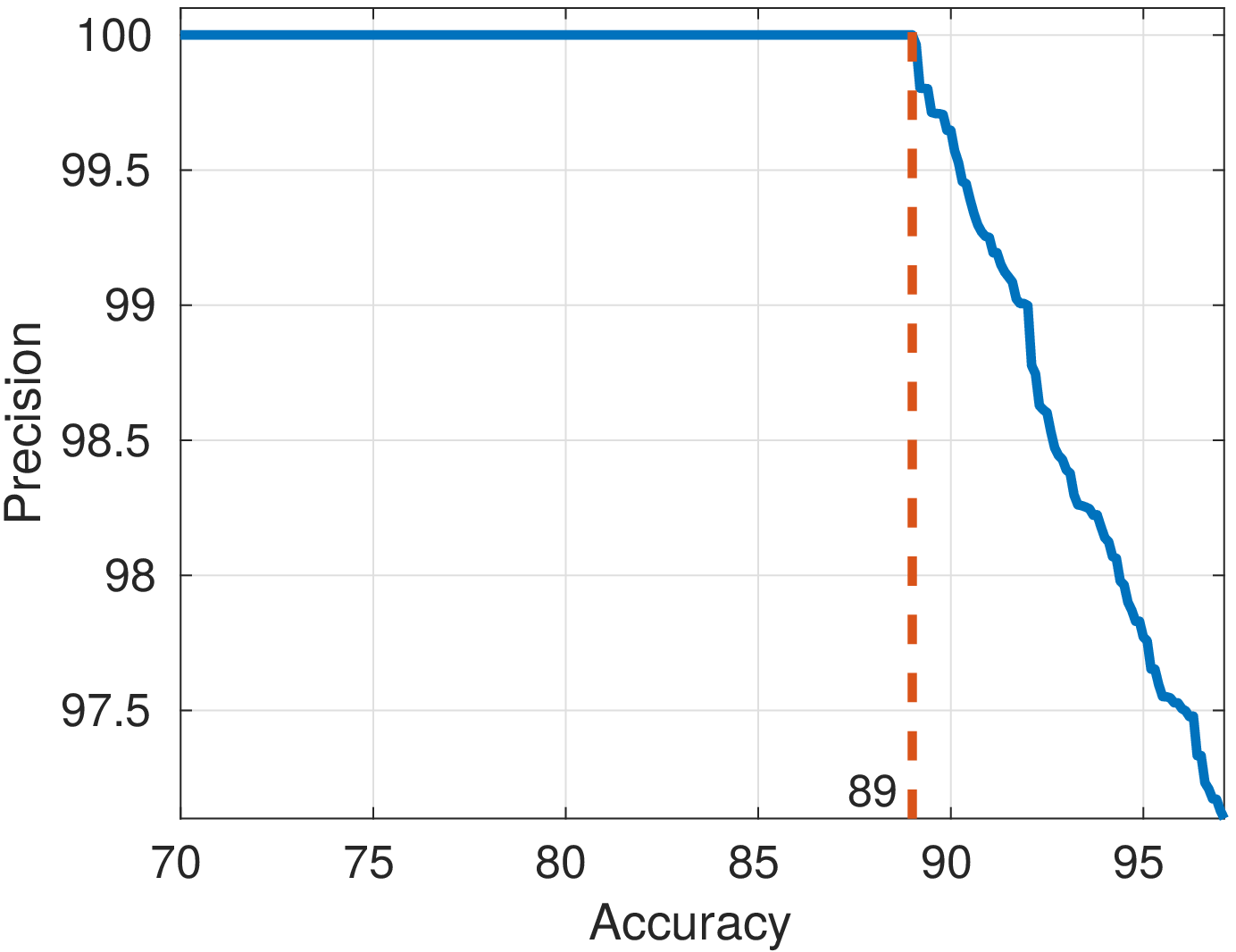}
      \caption{No occlusion}
      \label{fig:sub1}
    \end{subfigure}%
    \begin{subfigure}{.45\textwidth}
      \centering
      \includegraphics[width=1\linewidth]{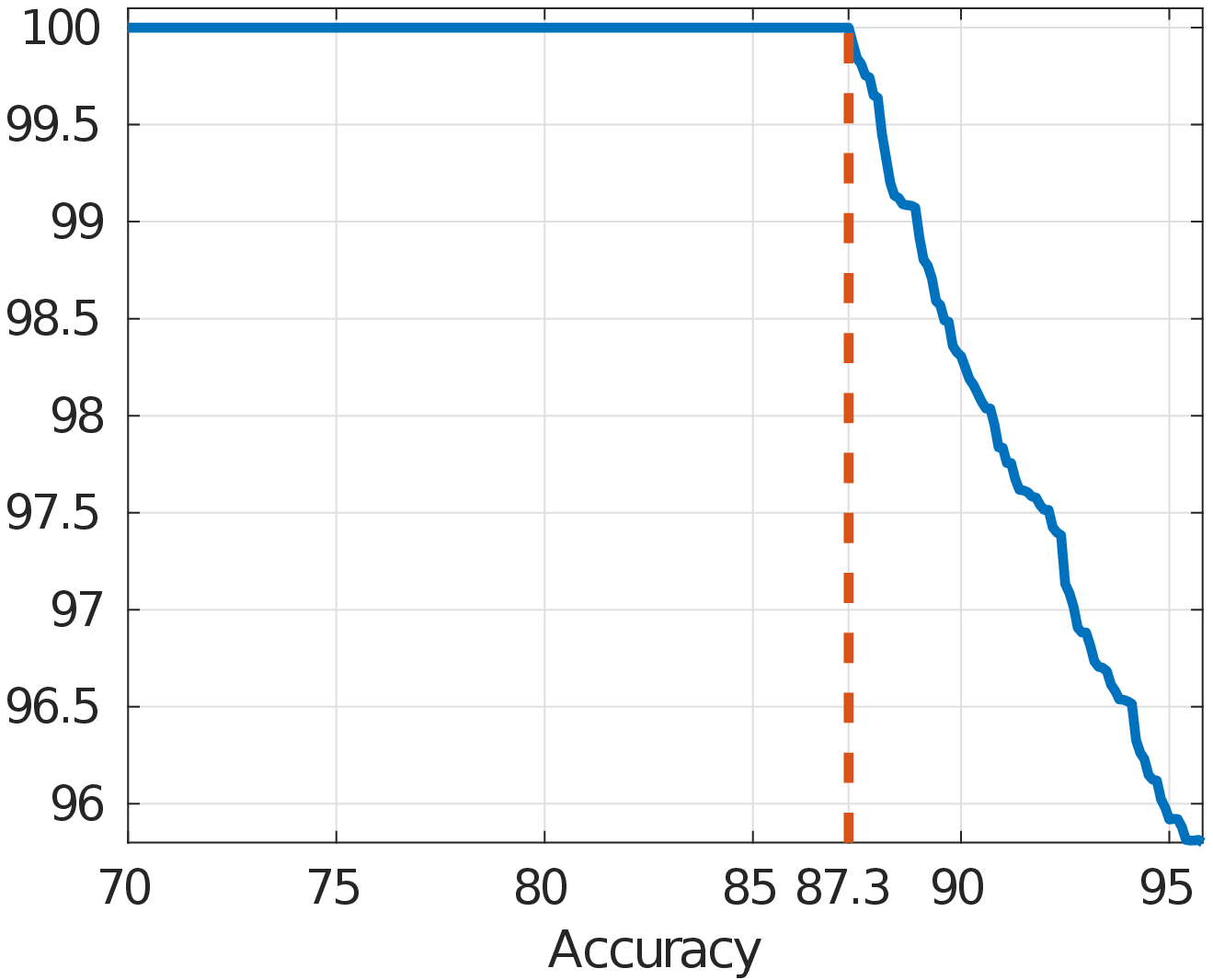}
      \caption{With occlusion}
      \label{fig:sub2}
    \end{subfigure}
    \caption{Plot of accuracy-precision curves for labelling frames without and with occlusions. In the latter case an average of 2.5 markers was missing in each frame.}
    \label{fig:Plots}
\end{figure}
Having defined the degree of confidence, there is the option to only assign a label to a marker if the corresponding degree of confidence is higher than a threshold. Otherwise, the marker is left unlabelled. This allows the model to set a trade-off between the precision (the fraction of correctly labelled markers over labelled markers) and accuracy (the fraction of correctly labelled markers over all markers). Fig. \ref{fig:Plots} shows the accuracy-precision curves.  It can be seen that a high proportion of the markers can be labelled with no error ($89\%$ when there is no occlusion and $87.3\%$ with an average of 2.5 markers occluded in each frame) and leaving less than $12.7 \%$ markers to be labelled manually.
\subsection{Trajectory Labelling Evaluation} \label{subsection: Trajectory Labelling Evaluation}
So far, we looked at frames individually. In the next stage, we integrate frame-by-frame labelling with information about the temporal order of the frames in an effort to label continuous trajectories. To evaluate the trajectory labelling stage, we used the original unlabelled test set and introduced occlusions to each motion sample with different occlusion ratios. Here, the occlusions are introduced as gaps with different lengths into trajectories. As a result, the motion of the marker is fragmented into two or more trajectories. 
Table \ref{tab:table2} shows the performance of trajectory labelling with different settings of $p$ and $q$ in our scoring function. When $p = 0$ and $q = 0$, this stage acts as a voting function ($S_{i}=|T_{ i }|$). Thus, the degree of confidence does not have an influence on the final result. For $p = 1$ and $q = 0$, the scoring function computes the sum of confidences for the frames that the label has been assigned to the marker. Also, when $p = 1$ and $q = -1$, the score for each label is considered as the average of confidences. Therefore, the number of times that a label is assigned to a marker is neutralized by averaging. Best performance in our hyper-parameters search was achieved when $p = 2$ and $q = -1/2$. In this case, the influence of $N_{i}$ on the score is less than when $q = -1$. Also, since $p$ is set to $2$, the role of the high degree of confidences is more than other settings.
\begin{table}[t]
    \centering
    \caption{The performance of trajectory labelling with different settings of $p$ and $q$ in our scoring function.}
    \label{tab:table2}
    \begin{tabular}{|c|c|c|c|c|c|c|}
    \hline
    \diagbox[ dir=NW,innerleftsep=.1cm,innerrightsep=5pt]{Method}{Occs Ratio} & 0 & 2\% & 4\% & 6\% & 8\% & 10\%\\
    \hline \hline
    No trajectory labelling &  97.13\% & {96.55}\% & {96.17}\%  & {95.91}\% & {95.68}\% & {94.82}\%\\ \hline
    $p = 0$, $q = 0$ & {97.52}\% & 96.81\% & 96.55\% & 96.09\% & 96.51\% & 95.43\% \\ \hline
    $p = 1$, $q = 0$ & 98.77\% & 98.01\% & 97.78\% & 97.32\% & 97.12\% & 96.85\% \\ \hline
    $p = 1$, $q = -1$ & 97.35\% & 96.51\% & 96.42\% & 96.06\% & 95.89\% & 94.37\%\\ \hline
    $p = 2$, $q = -1/2$ & \bfseries{99.85}\% & \bfseries{99.54}\% & \bfseries{99.47}\% & \bfseries{99.25}\% & \bfseries{99.07}\% & \bfseries{98.76}\%\\ \hline
    \end{tabular}
\end{table}
\section{Conclusion}
In this paper, we presented a method to address the problem of auto-labelling markers in optical motion capture pipelines. The essence of our approach was to frame the problem of single-frame labelling as a permutation learning task where the ordered set of markers can be recovered by estimating the permutation matrix from a shuffled set of markers. We exploited the idea of using DSMs to represent a distribution over the permutations. Also, we proposed a robust solution to correct the mislabelled markers by utilizing the temporal information where the label with a higher confidence score is assigned to the whole trajectory. We demonstrated that our method performed with very high accuracy even with only single-frame inputs, and when individual markers were occluded. Furthermore, the trajectory labelling will further improve if longer gap-free trajectories are available. Our method can be considered as both initialization and tracking. Once the model is trained, it easily runs on a medium-power CPU at $120$ frames per second and can therefore be used for real time tracking.
\section{Future Works}
Our model makes fast and robust predictions and is easy to train. However, it should be trained on a training set with the same marker layout. One solution could be to synthesize desired training sets by putting virtual markers on the animated body meshes from labelled data using body and motion animating tools such as \cite{Loper2014} and to record the motion of virtual markers. 

We have addressed the problem of single-subject marker labelling, but have not considered multi-subject scenarios. Future work could explore using clustering approaches and multi-hypothesis generative approaches to separate the subjects and apply the model on each of them.

Here, we have used a completely data-driven approach to label motion capture trajectories. The model could be further improved by integrating both anthropometric and kinematic information into our method where they can serve as priors that further constrain the model. 
%
%
%
\bibliographystyle{splncs04}
\bibliography{Autolabelling.bbl}

\begin{thebibliography}{10}
\providecommand{\url}[1]{\texttt{#1}}
\providecommand{\urlprefix}{URL }
\providecommand{\doi}[1]{https://doi.org/#1}

\bibitem{Adams2011}
{Adams}, R.P., {Zemel}, R.S.: {Ranking via Sinkhorn Propagation}. ArXiv pp.
  1106--1925 (2011)

\bibitem{Bergstra2012}
Bergstra, J., Bengio, Y.: Random search for hyper-parameter optimization.
  Journal of Machine Learning Research  \textbf{13},  281--305 (2012)

\bibitem{Birkhoff1946}
Birkhoff, G.: Three observations on linear algebra. Univ. Nac. Tacuman, Rev.
  Ser. A  \textbf{5},  147--151 (1946)

\bibitem{Etemad2016}
Etemad, S.A., Arya, A.: Expert-driven perceptual features for modeling style
  and affect in human motion. IEEE Transactions on Human-Machine Systems
  \textbf{46}(4),  534--545 (2016)

\bibitem{Han2018}
Han, S., Liu, B., Wang, R., Ye, Y., Twigg, C.D., Kin, K.: Online optical
  marker-based hand tracking with deep labels. ACM Transactions on Graphics
  \textbf{37}(4), ~166 (2018)

\bibitem{He2016}
He, K., Zhang, X., Ren, S., Sun, J.: Deep residual learning for image
  recognition. In: IEEE Conference on Computer Vision and Pattern Recognition
  (2016)

\bibitem{Herda2001}
Herda, L., Fua, P., Pl{\"{a}}nkers, R., Boulic, R., Thalmann, D.: {Using
  skeleton-based tracking to increase the reliability of optical motion
  capture}. Human Movement Science  \textbf{20}(3),  313--341 (2001).
  \doi{10.1016/S0167-9457(01)00050-1}

\bibitem{Holden2018}
Holden, D.: {Robust Solving of Optical Motion Capture Data by Denoising}. ACM
  Transactions on Graphics  \textbf{38}(1),  1--12 (2018).
  \doi{10.11499/sicejl1962.40.735}

\bibitem{Holzreiter2005}
Holzreiter, S.: Autolabeling 3d tracks using neural networks. Clinical
  Biomechanics  \textbf{20}(1), ~1--8 (2005)

\bibitem{Ionescu2014}
Ionescu, C., Papava, D., Olaru, V., Sminchisescu, C.: Human3.6m: Large scale
  datasets and predictive methods for 3d human sensing in natural environments.
  IEEE transactions on pattern analysis and machine intelligence
  \textbf{36}(7),  1325--1339 (2014)

\bibitem{Loper2014}
Loper, M., Mahmood, N., Black, M.J.: Mosh: Motion and shape capture from sparse
  markers. ACM Transactions on Graphics  \textbf{33}(6), ~220 (2014)

\bibitem{Maycock2015}
Maycock, J., R{\"{o}}hlig, T., Schr{\"{o}}der, M., Botsch, M., Ritter, H.:
  {Fully Automatic Optical Motion Tracking using an Inverse Kinematics
  Approach}. . In IEEE/RAS International Conference on Humanoid Robots pp.~2--7
  (2015). \doi{10.1109/HUMANOIDS.2015.7363590}

\bibitem{Meyer2014}
Meyer, J., Kuderer, M., Muller, J., Burgard, W.: {Online marker labeling for
  fully automatic skeleton Tracking in optical motion capture}. In Proc. IEEE
  International Conference on Robotics and Automation pp. 5652--5657 (2014).
  \doi{10.1109/ICRA.2014.6907690}

\bibitem{Pons2015}
Pons-Moll, G., Romero, J., Mahmood, N., Black, M.J.: Dyna: A model of dynamic
  human shape in motion. ACM Transactions on Graphics (TOG)  \textbf{34}(4),
  ~120 (2015)

\bibitem{rezatofighi2018}
Rezatofighi, S.H., Kaskman, R., Motlagh, F.T., Shi, Q., Cremers, D.,
  Leal-Taix{\'e}, L., Reid, I.: Deep perm-set net: learn to predict sets with
  unknown permutation and cardinality using deep neural networks. arXiv
  preprint arXiv:1805.00613  (2018)

\bibitem{santa2017}
Santa~Cruz, R., Fernando, B., Cherian, A., Gould, S.: Deeppermnet: Visual
  permutation learning. In: Proceedings of the IEEE Conference on Computer
  Vision and Pattern Recognition. pp. 3949--3957 (2017)

\bibitem{Schubert2015}
Schubert, T., Gkogkidis, A., Ball, T., Burgard, W.: Automatic initialization
  for skeleton tracking in optical motion capture. In: Proc. IEEE International
  Conference on Robotics and Automation. pp. 734--739. IEEE (2015)

\bibitem{Sigal2010}
Sigal, L., Balan, A.O., Black, M.J.: Humaneva: Synchronized video and motion
  capture dataset and baseline algorithm for evaluation of articulated human
  motion. International Journal of Computer Vision  \textbf{87}(1-2), ~4 (2010)

\bibitem{Sinkhorn1964}
Sinkhorn, R.: A relationship between arbitrary positive matrices and doubly
  stochastic matrices. The Annals of Mathematical Statistics  \textbf{35}(2),
  876--879 (1964)

\bibitem{Troje2008}
Troje, N.F.: Retrieving information from human movement patterns. In: Shipley
  TF, Zacks JM, editors. Understanding events: How humans see, represent, and
  act on events. New York: Oxford University  \textbf{1},  308--334 (2008)

\bibitem{Yu2007}
Yu, Q., Li, Q., Deng, Z.: {Online motion capture marker labeling for multiple
  interacting articulated targets}. Computer Graphics Forum  \textbf{26}(3),
  477--483 (2007). \doi{10.1111/j.1467-8659.2007.01070.x}

\end{thebibliography}

\end{document}